\documentclass{article}
\usepackage{spconf,amsmath,graphicx}
\usepackage{booktabs}
\usepackage{url}

\usepackage{enumitem}
\setlist{nosep, leftmargin=14pt}

\usepackage{mwe} % to get dummy images

\usepackage{amssymb}
\usepackage{hyperref}
\usepackage{booktabs}
\usepackage{array} % Für mehr Kontrolle über die Spaltenbreite
\usepackage{cleveref}
\usepackage{svg}

\title{Low-resource finetuning of foundation models beats \\ state-of-the-art in histopathology}
\name{Benedikt Roth \textsuperscript{1,2,*} \thanks{\textsuperscript{*} contributed equally to this work}, Valentin Koch \textsuperscript{1,2,*}, Sophia J. Wagner \textsuperscript{1,2,*},  Julia A. Schnabel \textsuperscript{1,2,3}}

\secondlinename{\textit{Carsten Marr \textsuperscript{2,$\dagger$} \thanks{\textsuperscript{$\dagger$} corresponding author, \href{mailto:carsten.marr@helmholtz-munich.de}{carsten.marr@helmholtz-munich.de}}, Tingying Peng \textsuperscript{2,$\dagger$}} \thanks{\textsuperscript{$\dagger$} corresponding author, \href{mailto:tingying.peng@helmholtz-munich.de}{tingying.peng@helmholtz-munich.de}}}

\address{\textsuperscript{1} School of Computation and Information Technology, Technical University of Munich, Germany \\
\textsuperscript{2} Helmholtz Zentrum München - German Research Center for Environmental Health, Germany \\
\textsuperscript{3} School of Biomedical Engineering and Imaging Sciences, King’s College London, UK
}

\begin{document}
%\ninept
%
\maketitle
\begin{abstract}
To handle the large scale of whole slide images in computational pathology, most approaches first tessellate the images into smaller patches, extract features from these patches, and finally aggregate the feature vectors with weakly-supervised learning. The performance of this workflow strongly depends on the quality of the extracted features. Recently, foundation models in computer vision showed that leveraging huge amounts of data through supervised or self-supervised learning improves feature quality and generalizability for a variety of tasks. In this study, we benchmark the most popular vision foundation models as feature extractors for histopathology data. We evaluate the models in two settings: slide-level classification and patch-level classification. We show that foundation models are a strong baseline. 
Our experiments demonstrate that by finetuning a foundation model on a single GPU for only two hours or three days depending on the dataset, we can match or outperform state-of-the-art feature extractors for computational pathology. These findings imply that even with little resources one can finetune a feature extractor tailored towards a specific downstream task and dataset. This is a considerable shift from the current state, where only few institutions with large amounts of resources and datasets are able to train a feature extractor. We publish all code used for training and evaluation as well as the finetuned models \footnote{\url{https://github.com/beneroth13/dinov2}}.
\end{abstract}
\section{Introduction}
\label{sec:intro}

\begin{figure}[t]
  \centering
  \includegraphics[width=1.\linewidth]{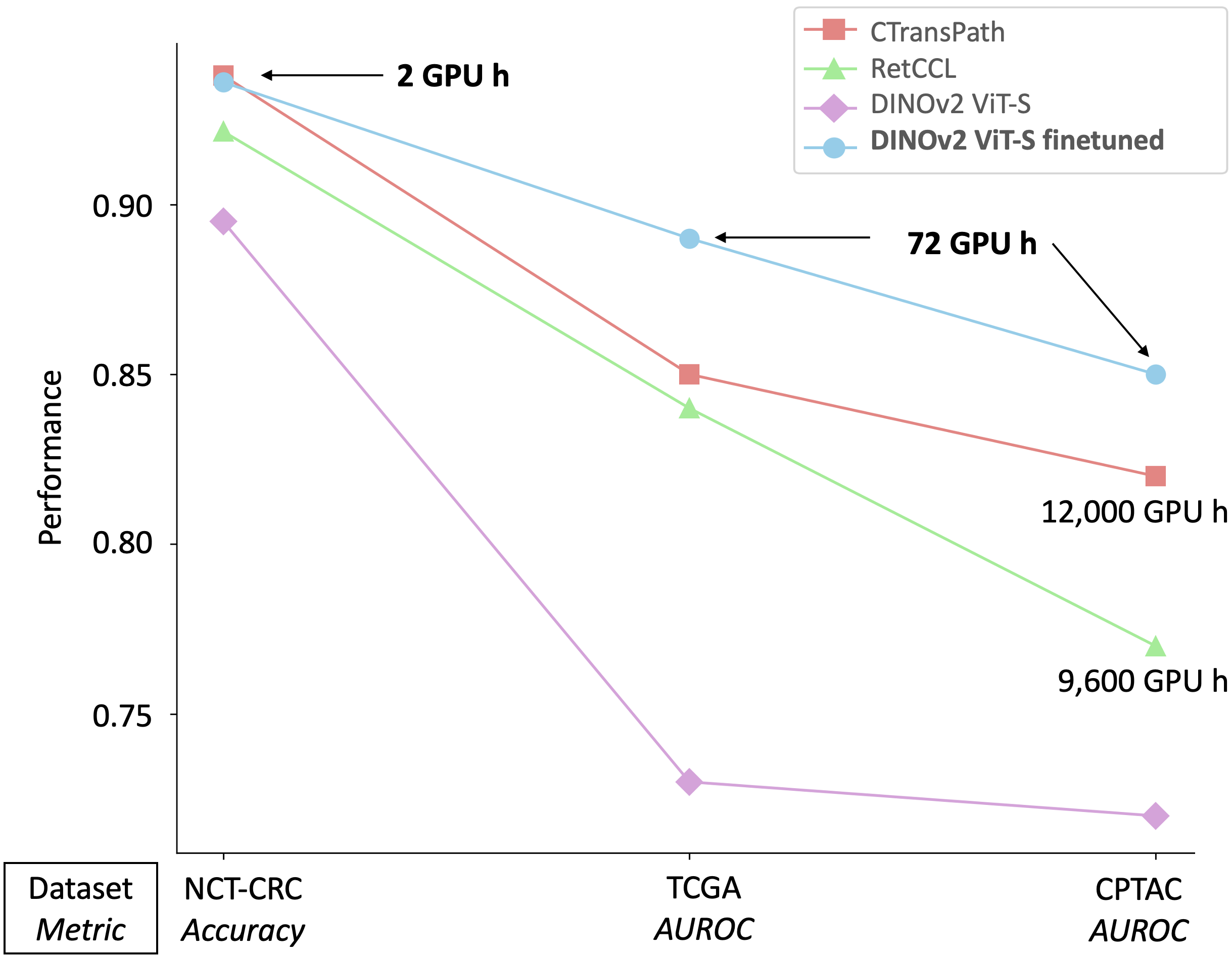}
  % \includesvg[width=1.\linewidth]{learning_curves/overview.svg}
  \caption{We propose finetuning a DINOv2 ViT-S, which yields at least equal performance compared to CTransPath and RetCCL but in a fraction of domain specific training time. Performance is measured on three datasets: TCGA \& CPTAC (WSI-level classification) and NCT-CRC (patch-level classification).}  
  \label{fig1}
\end{figure}

% \begin{itemize}
%     \item foundation models in LLM
%     \item Foundation models in CV
%     \item Lack of data in medical imaging
%     \item in computational pathology, recent trend of foundation models but all private and not accessible
%     \item our contributions:
%     \item benchmarked CV foundation models
%     \item finetuned best performing method on pathology tasks
% \end{itemize}

Recent progress in natural language processing, in particular the success of large language models~\cite{llmsurvey}, has shown that huge data sources can be leveraged to build models that generalize well to a wide variety of tasks. These models, also called foundation models \cite{bommasani2022opportunities}, are trained on large-scale datasets without task-specific supervision and can be adapted for specialized downstream applications. 

Similar approaches have been developed for computer vision, where the most influential aspects for a good performance lie in large-scale datasets and self-supervised learning techniques. The core concept is to efficiently represent complex visual data in a lower-dimensional space. One common approach is contrastive learning, which involves pushing similar instances closer in the learned representation space while separating dissimilar ones~\cite{chen2020simple}. Training networks with contrastive loss often needs a large batch size and has the risk to collapse to one representation. To account for this, non-contrastive methods, such as MoCov3 \cite{chen2021empirical}, were proposed. MoCov3 creates a dynamic dictionary of negative examples using a momentum update mechanism. A similar approach is student-teacher network training, where the student only gets a masked version of the input and tries to learn similar representations as the teacher network. The student is updated during training whereas the teacher is updated with the exponential moving average of the student weights~\cite{hu2022teacherstudent}. This strategy can be used for knowledge distillation, where the knowledge from a larger model is transferred to a smaller one. 

Translating these efforts to medical tasks would be highly beneficial but is challenging for several reasons: Large-scale and publicly available data sources are scarce, imaging modalities and downstream tasks are very diverse, high performance on very specific tasks is preferred over a good performance on a broad set of tasks, and, finally, the patients' privacy needs to be preserved. Nevertheless, a few models were proposed as potential foundation models that are trained on medical imaging data, but not extensively on histopathology data~\cite{remedis, moor2023medflamingo}.

In computational pathology, self-supervised feature extractors are commonly used to compute a low-dimensional representation of a whole slide image that was tessellated into smaller patches~\cite{kimianet,retccl,Ctranspath}. These low-dimensional features are then usually aggregated for a downstream task using weakly-supervised learning. The largest dataset used for pre-training these feature extractors comprises 15 million patches~\cite{Ctranspath}. However, this is still magnitudes smaller than the large-scale pre-training datasets used in classical computer vision, e.g., 142 million images curated from a dataset of 1.2 billion images for DINOv2~\cite{dinov2}. There has been a recent effort to scale-up the pre-training for histopathology-specific feature extractors using private datasets~\cite{Dinov2_UNI,vorontsov2023virchow,campanella2023computational} but none of the models is currently publicly available and could be included in this study. 

Chen et al.~\cite{Dinov2_UNI} report using 32 NVIDIA A100 80GB GPUs and a batch size of 3,072 for 125,000 iterations without specifying the exact training time. CTransPath was trained for 250 hours on 48 NVIDIA V100 GPUs~\cite{Ctranspath}. RetCCL~\cite{retccl} used 32 NVIDIA V100 32G GPU for 300 hours. Virchow ~\cite{vorontsov2023virchow} and REMEDIS ~\cite{remedis} do not specify their setup but the scale of the data and the trained model suggest a similar compute heavy setup. As the compute resources needed to train such a model is out of reach for many institutions, only publicly available models such as CTransPath ~\cite{Ctranspath} or RetCCL ~\cite{retccl} are used. 

In our work, we propose a shift towards low-resource fine-tuning on specific datasets and show its applicability. We present a benchmark of four of the largest foundation models in classical computer vision on histopathology downstream tasks and compare the models to the state-of-the-art domain-specific feature extractors CTransPath and RetCCL. We finetune ViT models of the best performing self-supervised learning method DINOv2 on a single GPU and show that finetuning foundation models on task-specific data can outperform large-scale trained domain-specific models while needing only a fraction of resources and training time (\Cref{fig1}). 

\section{Experiments \& Methods}
\label{sec:experiments}

Our comparative analysis includes ResNet50~\cite{resnet}, pre-trained on ImageNet and its truncated variant as baseline models. We benchmark four vision foundation models that have been trained on natural images:
ImageBind, a multi-modal model trained on paired data consisting of images and a corresponding second modality such as text or depth-maps~\cite{imagebind};
Segment Anything (SAM), a segmentation model, trained with supervision on a large amount of images with segmentation masks~\cite{SAM};
BEiT, a vision transformer (ViT) trained with self-supervision based on masked image modeling~\cite{bao2022beit}; %A visual token is created using VAE, which is supposed to be recreated using the ViT based on the input including masked patches. 
DINOv2, a self-supervised teacher-student model trained on a large curated dataset~\cite{dinov2}. As histopathology-specific feature extractors, we evaluate the state-of-the-art models CTransPath~\cite{Ctranspath} and RetCCL~\cite{retccl}, which are specifically designed for patches from whole-slide images and are trained on 15 million histopathology patches. 

\subsection{Experimental setup}
We evaluate the methods on three datasets of colorectal cancer tissue, two with annotations for slide-level classification and one with patch-level annotations.

\textbf{Slide-level classification.}
The public repository The Cancer Genome Atlas (TCGA)~\cite{TCGA} contains whole-slide images of 632 patients with colorectal cancer (CRC) from the cohorts COAD and READ with annotations of microsatellite instability (MSI). We perform 5-fold cross validation using  three folds for training, one for validation, and one for testing. As an external test set, we use the CRC cohort of the public database The Clinical Proteomic Tumor Analysis Consortium, (CPTAC)~\cite{cptac} with 110 patients. 

We tessellate the whole slide images at 10$\times$ magnification in patches of size $256 \times 256$ pixels (px) and extract the features\footnote{Code for the pre-processing pipeline available at \url{https://github.com/peng-lab/HistoBistro/tree/feature_extraction}.}. Subsequently, we train a slide-level classifier using transformer-based aggregation following Wagner et al.~\cite{WAGNER20231650}.

\textbf{Patch-level classification.} The NCT-CRC-100K (NCT) dataset~\cite{nct-crc} contains 100,000 non-overlapping image patches from H\&E-stained WSIs from nine classes of colorectal cancer and normal colon tissue. The images are of size $224 \times 224$ pixels (px) at 0.5 $\mu$m/px magnification and are color-normalized using Macenko’s method~\cite{5193250}. We use the full dataset for training. For testing, we use CRC-VAL-HE-7K~\cite{nct-crc}, an independent dataset of 7,180 patches from an external cohort with the same tissue classes.

We extract the features using the introduced models as feature extractors and evaluate the classification using the sklearn $K$-nearest neighbors implementation with $K = 20$ as well as the sklearn implementation of logistic regression classifier with 1,000 iterations and $\ell_2$-regularization coefficient $\lambda$ of $\frac{100}{MC}$, where $M$ is the embedding dimension and $C$ is the number of classes. The choice of hyperparameters and evaluation methods follows Chen et al.~\cite{chen2023generalpurpose}.

\begin{table}[t]
  \centering
  \begin{tabular}{lccc}
    \toprule
        & & \multicolumn{2}{c}{AUROC}   \\ \cmidrule(lr){3-4}
      Model  & Histo & TCGA & CPTAC\\
    \midrule
      ResNet50              &               & $0.67 {\scriptstyle \pm 0.10}$ & $0.65 {\scriptstyle \pm 0.05}$ \\
      ResNet50 truncated    &               & $0.68 {\scriptstyle \pm 0.02}$ & $0.68 {\scriptstyle \pm 0.02}$ \\
      SAM (ViT-B)           &               & $0.56 {\scriptstyle \pm 0.07}$ & $0.62 {\scriptstyle \pm 0.03}$ \\
      SAM (ViT-H)           &               & $0.55 {\scriptstyle \pm 0.05}$ & $0.64 {\scriptstyle \pm 0.02}$ \\
      BEiT (ViT-B)          &               & $0.52 {\scriptstyle \pm 0.07}$ & $0.51 {\scriptstyle \pm 0.02}$ \\
      ImageBind (ViT-h)     &               & $0.64 {\scriptstyle \pm 0.06}$  & $0.71 {\scriptstyle \pm 0.02}$ \\
      DINOv2 (ViT-S)        &               & $0.73 {\scriptstyle \pm 0.07}$ & $0.72 {\scriptstyle \pm 0.06}$ \\
      DINOv2 (ViT-g)        &               & $0.66 {\scriptstyle \pm 0.12}$ & $0.60 {\scriptstyle \pm 0.03}$  \\ 
      RetCCL (ResNet50)   & $\checkmark$  & $0.84 {\scriptstyle \pm 0.08}$ & $0.77 {\scriptstyle \pm 0.05}$\\
      CTransPath (Swin-T)   & $\checkmark$  & $\underline{0.85} {\scriptstyle \pm 0.05}$ & $\underline{0.82} {\scriptstyle \pm 0.04}$\\\hline
      DINOv2 (ViT-S)        & $\checkmark$  & $\textbf{0.89}{\scriptstyle \pm  0.05}$ & $\textbf{0.85} {\scriptstyle \pm 0.02}$ \\
      DINOv2 (ViT-g)        & $\checkmark$  & $0.84 {\scriptstyle \pm 0.05}$ & $0.79 {\scriptstyle \pm 0.03}$\\
    \bottomrule
  \end{tabular}
    \caption{Finetuned DINOv2 outperforms dedicated feature extractors CTransPath and RetCCL on AUROC scores on WSI classification (MSI detection) in CRC tissue from TCGA and CPTAC (external). Also, DINOv2 ViT-S notably outperforms the ViT-g variant.}
    \label{tab:auroc_TCGA}

  %\\[1.5pt] %You can adjust how far below the table the text should appear
\end{table}

\begin{table}[tpb]
  \centering
  %\begin{tabular} {p{2.7cm} *{1}{>{\centering\arraybackslash}p{0.4cm}} *{4}{>{\centering\arraybackslash}p{0.6cm}}}
  \begin{tabular} {lccccc}
    \toprule
     & & \multicolumn{2}{c}{20-NN} & \multicolumn{2}{c}{linear probe} \\ \cmidrule(lr){3-4} \cmidrule(lr){5-6}
      Model & Histo & ACC & F1 & ACC & F1\\
    \midrule
      ResNet50          & & 0.78 & 0.81 & 0.86 & 0.88\\
      ResNet50 truncated & & 0.87 & 0.91 & 0.88 & 0.89\\
      SAM (ViT-B)              & & 0.75 & 0.76 & 0.83 & 0.81\\
      SAM (ViT-H)              & & 0.69 & 0.71 & 0.79 & 0.84\\
      BEiT (ViT-B)              & & 0.63 & 0.66 & 0.59 & 0.65 \\
      ImageBind (ViT-h)         & & 0.87 & 0.89 & 0.90 & 0.91\\
      DINOv2 (ViT-S)      & & 0.88 & 0.90 & 0.90 & 0.92\\
%      DINOv2 ViT-B & 0.882 & 0.896 & 0.89 & 0.918\\
%      DINOv2 ViT-L & 0.887 & 0.905 & 0.908 & 0.933\\
      DINOv2 (ViT-g)      & & 0.91 & 0.93 & 0.92 & 0.94\\
      RetCCL (ResNet50)   & $\checkmark$ & 0.91 & 0.93 & 0.92 & 0.94\\
      CTransPath (Swin-T) & $\checkmark$ &\textbf{0.95} & \textbf{0.96} & \textbf{0.94} & \underline{0.95}\\
      \hline
      DINOv2 (ViT-S) &  $\checkmark$ & \underline{0.94} & \underline{0.95} & \underline{0.93}& 0.94 \\
      DINOv2 (ViT-g)  & $\checkmark$ & 0.93 & \underline{0.95} & \textbf{0.94} & \textbf{0.96} \\
    \bottomrule
  \end{tabular}
    \caption{Finetuned DINOv2 matches the performance of the best-scoring feature extractor, CTransPath, on balanced accuracy (ACC) and weighted F1-scores (F1) for CRC tissue classification on NCT-patches.}
      \label{tab:crc}

  %\\[1.5pt] %You can adjust how far below the table the text should appear
  \raggedright %
\end{table}

\subsection{Finetuning and implementation details}

To evaluate performance improvements of the foundation models by finetuning, we focus on DINOv2 in the smallest and largest variants ViT-S/14 (21M parameters, \Cref{tab:parameter}) and ViT-g/14 (1.1B parameters), respectively. We finetune both models using the original DINOv2 implementation\footnote{\url{https://github.com/facebookresearch/dinov2}} on the NCT and TCGA dataset. As a starting point, the publicly available pretrained weights for the backbones were loaded into both the student and the teacher network. Since the weights for the DINO-heads are not publicly available, a random initialization is used. As an additional augmentation we add vertical flipping.

Similar as proposed in~\cite{dinov2}, we use input images of size $224 \times 224$ px with a local crop size of $98 \times 98$ px. In order to adapt the positional encoding from its original size of $518 \times 518$ px to the desired size of $224 \times 224$ px, we employ bicubic interpolation. 
%Alternatively, we evaluate the models using a center crop of the positional embeddings, which results in slightly worse performance.

In our finetuning experiments, we utilize the provided configuration file from the training pipeline with a base learning rate of $2 \times 10^{-4}$. To account for the comparatively small datasets, we adjust the number of iterations per pseudo-epoch and the number of pseudo-epochs. As batch-size we use 256 for ViT-S and 32 for ViT-g. Using a smaller batch size of 128 for ViT-S yielded similar results as with batch size 256. In the KoLeoLoss the hyperparameter $\epsilon$ was changed from $1 \times 10^{-8}$ to $1 \times 10^{-4}$ to avoid infinite values in the loss. The remaining hyperparameters are kept consistent with the values specified in the configuration file provided in the official implementation. Training was done on a single NVIDIA A100 GPU (80GB).
%which is downscaled according to the batch size (bs) and number of GPUs ($N$), multiplying the base learning rate with $\sqrt{\frac{\mathrm{bs} \cdot \mathrm{N}}{1024}}$. The batch size was set to $256$ for ViT-S and to $32$ for ViT-g and number of GPUS $N=1$. (TODO: is there any adjustment from our side? if not we can remove probably)

\textbf{Slide-level classification.} For finetuning on TCGA, we tessellate the whole slide image into patches of size $512 \times 512$ px at $20\times$ magnification. We thereby obtain a training dataset of 2.5M patches, which are downsampled using RandomResizeCrop to $224 \times 224$ px as input to the vision transformer. We set the iterations per epoch to $500$ and the number of pseudo-epochs to $200$.
 
\textbf{Patch-level classification.}
We use the full training dataset NCT of 100,000 images for finetuning and set the iterations per epoch to $100$ and the number of pseudo-epochs to $100$, respectively.
%Our initial experiments were conducted using the NCT-CRC-100K dataset, where we employed approximately 400 iterations $\left( \frac{100000}{256} \right) $ while keeping the suggested 100 pseudo-epochs from the config-file. However, this strategy led to checkpoints achieving peak performance well before reaching 10,000 iterations, followed by a substantial decline in performance. To address this, we adjusted the length of each pseudo-epoch to 100, allowing training to conclude around the point of previously identified peak performance. Building upon the insights gained from the NCT-CRC-100K experiments, we applied a refined approach to the TCGA dataset. Given its larger size (25 times that of NCT-CRC-100K), we tried out 200 pseudo-epochs and 500 iterations, totaling 100,000 iterations. Although the performance plateaued around 30,000 to 40,000 iterations, we observed no significant decrease in performance. Thus, we continued the training for the full 100,000 iterations hoping for a late performance peak. 

% Maybe to discussion? As dino-head weights were missing, we confirmed we were finetuning and not training from scratch by comparing runs with pretrained backbone to random initialization. Comparisons of the loss curves confirmed that we were indeed finetuning.

\subsection{Results}

\begin{figure*}[htpb]
  \centering
  \includegraphics[width=1.0\linewidth]{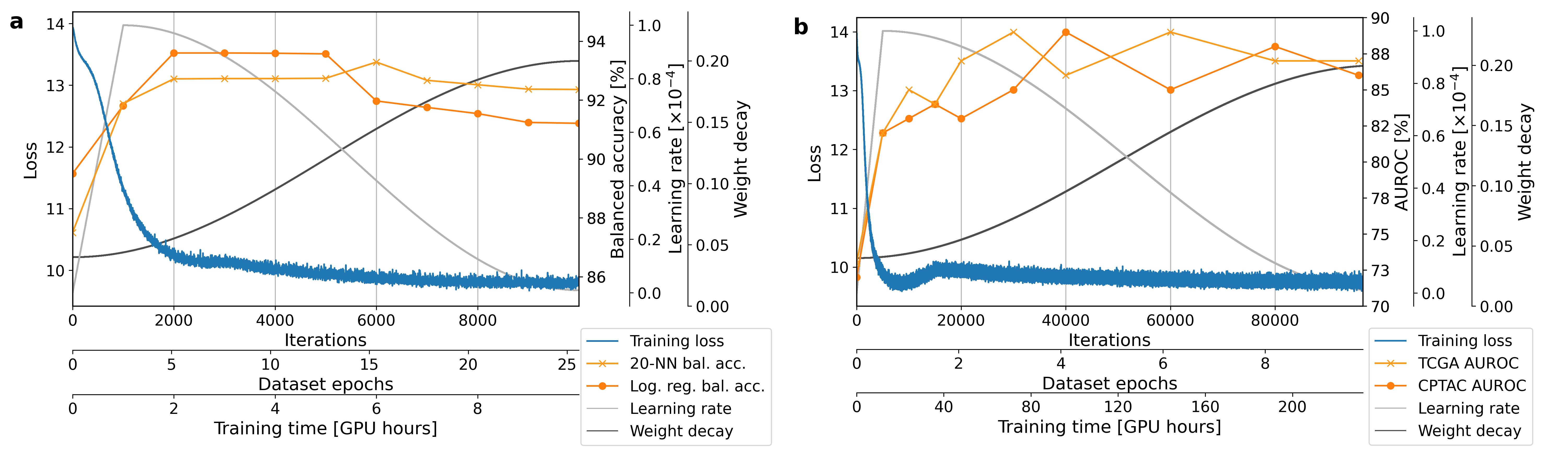}
  \caption{Performance over time of finetuning a ViT-s with DINOv2: a) on NCT-CRC and evaluating on the external NCT-CRC testset on patch-level classification and b) on TCGA and testing on TCGA (5-fold cross-validation) and CPTAC (external testset) on WSI-level classification.}
  \label{fig2}
\end{figure*}

\textbf{Slide-level classification.}
In our experiments, DINOv2 outperforms all other computer vision foundation models, i.e., SAM, BEiT, and ImageBind, in both the small (ViT-S) and giant (ViT-g) variant (\Cref{tab:auroc_TCGA}). Notably, the giant variant performs substantially worse than the small variant (e.g., 0.66 vs.\ 0.73 on TCGA, 0.6 vs. 0.72 on CPTAC). The performance of the larger network is surpassed by the truncated version of ResNet50 trained on ImageNet, which consistently outperforms its full counterpart. Interestingly, the features from ImageBind lead to a strong generalization on CPTAC, where the external validation almost matches the performance of DINOv2 (0.71 vs. 0.72) despite a low performance on the in-domain test set TCGA (0.64). These models are all outperformed by the histopathology feature extractors, RetCCL and CTransPath, showing that feature quality benefits from training on histopathology-specific data. However, finetuning DINOv2 on our 2.5M patches from TCGA further improves the performance and outperforms the state-of-the-art feature extractors, where the smaller variant based on ViT-S performs better than the large version. Compared to CTransPath we only utilize 0.6\% (72 h vs.\ 12,000 h) of GPU hours to reach peak performance for training the ViT-S models (Figure \ref{fig1}). Peak performance is reached at approximately 30,000 to 40,000 iterations for ViT-S compared to around 60,000 for ViT-g (\Cref{fig2}).

\textbf{Patch-level classification.}
As for slide-level classification, DINOv2 outperforms all other computer vision foundation models, where the margin to ImageBind and the truncated ResNet50 version is small. Histopathology-specific pre-training improves also in this case, and CTransPath achieves best results accross all metrics (\Cref{tab:crc}). However, RetCLL obtains an almost identical performance. Finetuning both DINOv2 variants also almost match the performance of CTransPath with only two GPU hours (0.08\% of CTransPath's) of training (\Cref{fig1}). The peak performance is approximately reached at 2,000 iterations for ViT-S compared to around 10,000 for ViT-g (\Cref{fig2}).

In both tasks, all foundation models show a very similar behavior: ImageBind and the truncated ResNet50 follow DINOv2 regarding the performance, where ImageBind has roughly 20 times more parameters (\Cref{tab:parameter}). Conversely, both variants of SAM as well as BEiT underperform compared to other models. The training duration for NCT is significantly shorter than for TCGA due to the smaller number of slides in the training set. To ensure no overfitting is happening, we evaluate each task on an external test set, CPTAC and CRC-VAL-HE-7K, repsecitvely, as well as the in-domain TCGA test set.

\begin{table}[htpb]
  \centering
  \begin{tabular} {l r r}
    \toprule
      Model & \#params & feature size\\
    \midrule
      ResNet50          & 25.6 M & 1024 \\
      ResNet50 truncated & 8.5 M & 1024 \\
      SAM (ViT-B)       & 91 M & 256 \\
      SAM (ViT-H)       & 636 M & 256 \\
      BEiT (ViT-B)      & 85.7 M & 768 \\
      ImageBind (ViT-h) & 630 M & 1024 \\
      RetCCL (ResNet50) & 23.5 M & 2048 \\
      CTransPath (Swin-T) & 27.5 M & 768 \\
      DINOv2 (ViT-S)    & 21 M & 384 \\
      DINOv2 (ViT-g)    & 1.1 B & 1536 \\
    \bottomrule
  \end{tabular}
    \caption{Parameter count and feature dimension of benchmarked models.}
  \label{tab:parameter}
%  \\[1.5pt] %You can adjust how far below the table the text should appear
\end{table}

\section{Conclusion}
\label{sec:conclusion}
 %Recently, histopathology foundation models trained on millions of whole-slide image evolved \cite{Dinov2_UNI, vorontsov2023virchow, campanella2023computational}. 

In our study, we benchmarked four of the most popular publicly available state-of-the-art foundations models in the field of computer vision as feature extractors. By fine-tuning the most promising model with the DINOv2 framework, we achieved notable improvements that are comparable or better than the current state-of-the-art feature extractor in histopathology CTransPath, while only using $0.08\%$ or $0.6\%$ of their compute budget, which amounts to two hours or three days of training on one single A100 GPU depending on the dataset.
% Moreover, our experiments suggest that very small batch sizes such as 32 on ViT-g lead to considerable improvements, a counter-intuitive finding, as it was often shown that for self-supervision large batch sizes are beneficial \cite{grill2020bootstrap}. Experiments for fintuning with smaller batch sizes of 128 on ViT-S do not show any performance drops compared to batch size 256. \\ 
%\textbf{Limitations}
%It is important to point out the limitations of this study as well.
Our experiments were conducted with a limited scope, utilizing two datasets for training and three for testing purposes. To fully conclude a superior performance compared to established feature extractors, more experiments are needed on more diverse benchmarks. 

\section{Ethics statement}
\label{sec:ethics}

This research study was conducted retrospectively using human subject data made available in open access. Ethical approval was not required as confirmed by the license attached with the open access data.

\section{Acknowledgments}
\label{sec:acknowledgments}
V.K. and S.J.W. were supported by the Helmholtz Association under the joint research school “Munich School for Data Science - MUDS”. S.J.W. was supported by the Add-on Fellowship of the Joachim Herz Foundation. This work was also supported by the BMBF-funded de.NBI Cloud within the German Network for Bioinformatics Infrastructure (de.NBI) (031A532B, 031A533A, 031A533B, 031A534A, 031A535A, 031A537A, 031A537B, 031A537C, 031A537D, 031A538A). C.M. has received funding from the European Research Council under the European Union’s Horizon 2020 research and innovation program (grant agreement number 866411) and is supported by the Hightech Agenda Bayern. The results shown here are in part based upon data generated by the TCGA Research Network \footnote{\url{https://www.cancer.gov/tcga}}.

\bibliographystyle{IEEEbib}
\bibliography{strings,refs}

\end{document}